%% file: maga.tex
\documentclass[conference]{IEEEtran}
\IEEEoverridecommandlockouts
\usepackage{cite}
\usepackage{amsmath,amssymb,amsfonts}
\usepackage{algorithmic}
\usepackage{graphicx}
\usepackage{textcomp}
\usepackage{xcolor}
\usepackage{subfigure}
\usepackage{booktabs}
\usepackage{multirow}
\def\BibTeX{{\rm B\kern-.05em{\sc i\kern-.025em b}\kern-.08em
    T\kern-.1667em\lower.7ex\hbox{E}\kern-.125emX}}
\begin{document}

\title{Multivariate Time-series Anomaly Detection via Graph Attention Network}

\author{\IEEEauthorblockN{Hang Zhao\IEEEauthorrefmark{1}\textsuperscript{\textsection}, Yujing Wang\IEEEauthorrefmark{1}\textsuperscript{\textsection}, Juanyong Duan\IEEEauthorrefmark{1}, Congrui Huang\IEEEauthorrefmark{1}, Defu Cao\IEEEauthorrefmark{1}, \\ Yunhai Tong\IEEEauthorrefmark{2}, Bixiong Xu\IEEEauthorrefmark{1}, Jing Bai\IEEEauthorrefmark{1}, Jie Tong\IEEEauthorrefmark{1}, Qi Zhang\IEEEauthorrefmark{1}}
\IEEEauthorblockA{\IEEEauthorrefmark{1}Microsoft, \IEEEauthorrefmark{2}Key Laboratory of Machine Perception, MOE, School of EECS, Peking University
}
\{hang.zhao, yujwang, juanyong.duan, conhua, t-decao, bix, jbai, jietong, qizhang\}@microsoft.com, yhtong@pku.edu.cn
}
    % \affiliation{ 
    %   \institution{Microsoft}
    %   \state{Beijing}
    %   \country{China}
    % }
    % \email{{yuyi,juaduan,chuwan,yujwang,conhua,bix}@microsoft.com}
    
% \author{\IEEEauthorblockN{Hang Zhao}
% \IEEEauthorblockA{\textit{Microsoft} \\
% hang.zhao@microsoft.com}
% \and
% \IEEEauthorblockN{Yujing Wang}
% \IEEEauthorblockA{\textit{Microsoft} \\
% yujwang@microsoft.com
% }
% \and
% \IEEEauthorblockN{Juanyong Duan}
% \IEEEauthorblockA{\textit{Microsoft} \\
% juanyong.duan@microsoft.com}
% \and
% \IEEEauthorblockN{Congrui Huang}
% \IEEEauthorblockA{\textit{Microsoft} \\
% conhua@microsoft.com}
% \and
% \IEEEauthorblockN{Defu Cao}
% \IEEEauthorblockA{\textit{Microsoft} \\
% t-decao@microsoft.com}
% \and
% \IEEEauthorblockN{Yunhai Tong}
% \IEEEauthorblockA{\textit{Peking University} \\
% yhtong@pku.edu.cn}
% \and
% \IEEEauthorblockN{Bixiong Xu}
% \IEEEauthorblockA{\textit{Microsoft} \\
% bix@microsoft.com}
% \and
% \IEEEauthorblockN{Jing Bai}
% \IEEEauthorblockA{\textit{Microsoft} \\
% jbai@microsoft.com}
% \and
% \IEEEauthorblockN{Jie Tong}
% \IEEEauthorblockA{\textit{Microsoft} \\
% jietong@microsoft.com}
% \and
% \IEEEauthorblockN{Qi Zhang}
% \IEEEauthorblockA{\textit{Microsoft} \\
% qizhang@microsoft.com}
% }

\maketitle
\begingroup\renewcommand\thefootnote{\textsection}
\footnotetext{Equal contribution}
\endgroup
\input{tex/abstract}
\begin{IEEEkeywords}
multivariate time-series, anomaly detection, graph attention network
\end{IEEEkeywords}
\input{tex/intro} 
\input{tex/related_work} 
\input{tex/methodology}

\input{tex/experiments}

\input{tex/qualitative_results.tex}

\input{tex/conclusion}

\bibliographystyle{IEEEtran}
\bibliography{IEEEexample}

\end{document}

%% file: tex/abstract.tex
\begin{abstract}
Anomaly detection on multivariate time-series is of great importance in both data mining research and industrial applications. Recent approaches have achieved significant progress in this topic, but there is remaining limitations. One major limitation is that they do not capture the relationships between different time-series explicitly, resulting in inevitable false alarms. In this paper, we propose a novel self-supervised framework for multivariate time-series anomaly detection to address this issue. Our framework considers each univariate time-series as an individual feature and includes two graph attention layers in parallel to learn the complex dependencies of multivariate time-series in both temporal and feature dimensions. In addition, our approach jointly optimizes a forecasting-based model and a reconstruction-based model, obtaining better time-series representations through a combination of single-timestamp prediction and reconstruction of the entire time-series. We demonstrate the efficacy of our model through extensive experiments. The proposed method outperforms other state-of-the-art models on three real-world datasets. Further analysis shows that our method has good interpretability and is useful for anomaly diagnosis.
%two major issues remain unsettled. many of previous methods pay much attention to temporal dependencies of multivariate time-series, but they neglect the importance of inter-feature relationships. Second, previous models can be classified as either a forecasting-based model that relies on single point prediction, or a reconstruction-based model learning a representation of the entire series. We show that there are situations that either model cannot deal with and they are actually complementary to each other.
% We propose Graph Attention Recurrent Network (GARN), a novel framework for multivariate anomaly detection. Our framework combines two key insights: (1) one can learn the temporal dependencies and correlations between different features in multivariate series simultaneously for performance boost and (2) forecasting-based and reconstruction-based models are complementary with each other in the multivariate anomaly detection task. Our framework includes two instances of graph attention network in parallel to learn the complex dependencies of multivariate series both in the temporal dimension and the feature-wise dimension. By jointly optimizing the loss functions of both forecasting-based and reconstruction-based models, GARN achieves superior performances on two public datasets. It also achieves nearly $10\%$ improvement on a large-scale industry dataset with 3 times inference speed faster. GARN also has a good interpretation accuracy of {\color{red} acc}.
\end{abstract}

%% file: tex/intro.tex
\section{Introduction}
\begin{figure}[htbp]
	\centering
	\includegraphics[width=0.48\textwidth]{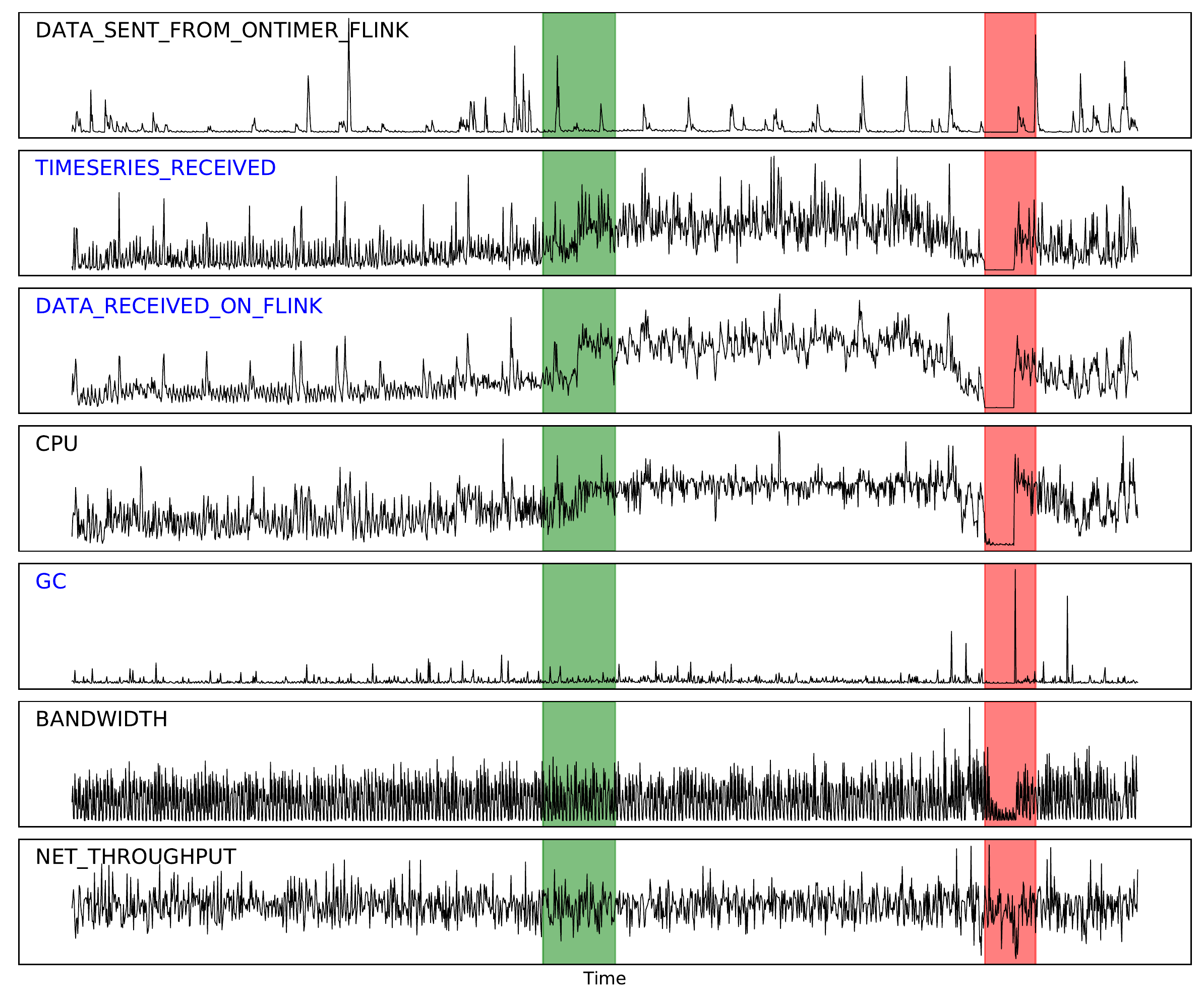}
	\caption{An example of multivariate time-series input. Green indicates normal values and red indicates anomalies.}
	%\caption{An example which may trigger false alarms on univariate anomaly detection models.}
	\label{fig:intro}
\end{figure}

% Time-series anomaly detection is an important research topic in data mining and has a wide range of applications in real life. It can be helpful on monitoring services and devices~\cite{su2019robust}. 
Time-series anomaly detection is an important research topic in data mining and has a wide range of applications in industry. Efficient and accurate anomaly detection helps companies to monitor their key metrics continuously and alert for potential incidents on time~\cite{ren2019time}. In real applications, multiple time-series metrics are collected to reflect the health status of a system~\cite{hundman2018detecting}. Univariate time-series anomaly detection algorithms are able to find anomalies for a single metric. However, it could be problematic in deciding whether the whole system is running normally. For example, sudden changes of a certain metric do not necessarily mean failures of the system. As shown in Figure \ref{fig:intro}, there are obvious boosts in the volumes of \textit{\footnotesize{TIMESERIES\_RECEIVED}} and \textit{\footnotesize{DATA\_RECEIVED\_ON\_FLINK}} in the green segment, but the system is still in a healthy state as these two features share consistent tendency. However, in the red segment, \textit{\footnotesize{GC}} shows inconsistent pattern with other metrics, indicating a problem in garbage collection. Consequently, \textit{it is essential to take the correlations between different time-series into consideration} in a multivariate time-series anomaly detection system. 

%Since those multiple metrics are highly correlated with each other, We choose an example from our online system for better understanding. Figure \ref{fig:intro} shows a case consisting of multiple metrics to indicate running state of the system. At the moment highlighted with green, the system is running normally, but obvious peaks can be detected in the metrics \textit{\footnotesize{TIMESERIES\_RECEIVED}} and \textit{\footnotesize{DATA\_RECEIVED\_ON\_FLINK}} (marked in blue). 
%In fact the two metrics represent input of the system in different modules. They are consistent with each other under normal circumstances. Moreover, anomalies have been found in the red part, as \textit{\footnotesize{GC}} shows inconsistent patterns with the input metrics which means problems in garbage collection have influenced the throughput of the system. Consequently multivariate time-series anomaly detection which takes \textit{the correlations between different time-series} into consideration is essential in monitoring a system.

Previous studies on multivariate time-series anomaly detection have made fruitful progresses. For instance, Malhotra et. al~\cite{malhotra2016lstm} proposes a LSTM-based encoder-decoder network that models reconstruction probabilities of the ``normal" time-series, and the reconstruction errors are utilized to detect anomalies from multiple sensors. Hundman et. al~\cite{hundman2018detecting} leverages LSTM to detect anomalies in multivariate time-series metrics of spacecraft based on prediction errors. OmniAnomaly~\cite{su2019robust} proposes a stochastic recurrent neural network, which captures the normal patterns of multivariate time-series by modeling data distribution through stochastic latent variables. However, to the best of our knowledge, none of previous works in the literature have addressed the problem of capturing multivariate correlations explicitly. \textit{We argue that there is still room for improvement if the relationships between different time-series can be modeled appropriately.}

\textit{\textbf{Our Solution:}} In this paper, we propose a novel framework --- MTAD-GAT (Multivariate Time-series Anomaly Detection via Graph Attention Network), to tackle the limitations of previous solutions. Our method considers each univariate time-series as an individual \textit{feature} and tries to model the correlations between different features explicitly, while the temporal dependencies within each time-series are modeled at the same time. The key ingredients in our model are two graph attention layers~\cite{velivckovic2017graph}, namely the \textit{feature-oriented graph attention} layer and the \textit{time-oriented graph attention} layer. The \textit{feature-oriented graph attention} layer captures the causal relationships between multiple features, and the \textit{time-oriented graph attention} layer underlines the dependencies along the temporal dimension. In addition, we jointly train a \textit{forecasting-based model} and a \textit{reconstruction-based model} for better representations of time-series data. The two models can be optimized simultaneously by a joint objective function.

The \textbf{contributions} of our paper are summarized as follows:
\begin{itemize}
\item We propose a novel framework to solve the multivariate time-series anomaly detection problem in a self-supervised manner. Our model shows superior performances on two public datasets and establishes state-of-the-art scores in the literature. It also achieves 9\% improvement for overall F1 score on our production data, bringing big impact on user satisfaction. 
\item For the first time, we leverage two parallel graph attention (GAT) layers to learn the relationships between different time-series and timestamps dynamically. Especially, our model captures the correlations between different time-series successfully without any prior knowledge.
%requires no prior knowledge about relations between features. Thorough analyses have shown the effectiveness of this design. 
% Without any given relations between features,  
%, which makes the model more robust and triggers less false alarms.
\item We integrate the advantages of both forecasting-based and reconstruction-based models by introducing a joint optimization target. The forecasting-based model focuses on single-timestamp prediction, while the reconstruction-based model learns a latent representation of the entire time-series. %They are complementary to each other by extensive experiments.
\item Our network has good interpretability. We analyze the attention scores of multiple time-series learned by the graph attention layers, and the results correspond reasonably well to human intuition. We also show its capability of anomaly diagnosis.
\end{itemize}

%% file: tex/related_work.tex
\section{Related Work}
There is a plenty of literature for time-series anomaly detection, which can be classified into two categories. The first category of approaches analyzes each individual time-series by applying univariate models~\cite{wong2015rad,kejariwal2015introducing,malhotra2015long,ren2019time}, while the second one models multiple time-series as a unified entity~\cite{hundman2018detecting,zong2018deep,li2018anomaly,li2019mad,park2018multimodal,su2019robust,mirsky2018kitsune}. From another perspective, existing anomaly detection models can also be categorized into two paradigms, namely \textit{forecasting-based} models~\cite{ding2018multivariate, hundman2018detecting,gugulothu2018sparse} and \textit{reconstruction-based} models~\cite{mirsky2018kitsune, li2019mad, su2019robust,park2018multimodal}. In this section, we summarize important works about time-series anomaly detection and discuss these two paradigms in detail.

\subsection{Univariate Anomaly Detection}
Classic methods typically utilize handcrafted features to model normal/anomaly event patterns~\cite{mousavi2015analyzing}, such as hypothesis testing~\cite{rosner1983percentage}, wavelet analysis~\cite{lu2008network}, SVD~\cite{mahimkar2011rapid} and ARIMA~\cite{zhang2005network}. Recently, Netflix has released a scalable anomaly detection solution based on robust principal component analysis~\cite{wong2015rad}, which has been proven successful in some real scenarios. Twitter has also published a seasonality-considered anomaly detection method using the Seasonal Hybrid Extreme Study Deviation test (S-H-ESD)~\cite{kejariwal2015introducing}. Recent advances in neural networks also lay a strong foundation for time-series anomaly detection~\cite{malhotra2015long, xu2018unsupervised, ren2019time}. DONUT~\cite{xu2018unsupervised} is an unsupervised anomaly detection method based on Variational Auto-Encoder (VAE), and SR-CNN~\cite{ren2019time} combines the benefits of Spectral Residual (SR) and convolutional neural network to achieve state-of-the-art performance on univariate time-series anomaly detection.

\subsection{Multivariate Anomaly Detection}

\subsubsection{Forecasting-based Models}
A forecasting-based model detects anomalies based on prediction errors~\cite{laptev2015generic}. %Forecasting-based models have been proven effective for time-series anomaly detection~\cite{hayton2006static, chauhan2015anomaly}. 
LSTM-NDT~\cite{hundman2018detecting} proposes an unsupervised and non-parametric thresholding approach to interpret predictions generated by an LSTM network. It builds up an automatic anomaly detection system to monitor the telemetry data sent back by the spacecraft. Ding et. al~\cite{ding2018multivariate} proposes a real-time anomaly detection algorithm based on Hierarchical Temporal Memory (HTM) and Bayesian Network (BN). Gugulothu et. al~\cite{gugulothu2018sparse} combines non-temporal dimensional reduction techniques and recurrent auto-encoders for time-series modeling through an end-to-end learning framework. DAGMM ~\cite{zong2018deep} focuses on anomaly detection of multivariate data without temporal dependencies. The input of DAGMM is just single entity observation (with multiple feature dimensions) instead of a temporal sequence. 

\subsubsection{Reconstruction-based models}
A reconstruction-based model learns the representation for the entire time-series by reconstructing the original input based on some latent variables. %, such as Denoising Autoencoders~\cite{sakurada2014anomaly} and Deep Belief Nets~\cite{wulsin2010semi}. 
Pankaj et. al~\cite{malhotra2016lstm} proposes an LSTM-based Encoder-Decoder framework to learn representations over normal time-series for anomaly detection. Kitsune~\cite{mirsky2018kitsune} is an unsupervised model, mapping the features of an instance to integrated visible neurons which are then used to reconstruct the features back by an autoencoder. Generative Adversarial Networks (GANs) have also been widely used in multivariate time-series anomaly detection. Instead of treating each time-series independently, MAD-GAN~\cite{li2019mad} considers the entire variable set concurrently to capture the latent interactions among variables. GAN-Li ~\cite{li2018anomaly} proposes a novel GAN-based anomaly detection method which deploys the GAN-trained discriminator together with the residuals between generator-reconstructed data and the actual samples. LSTM-VAE~\cite{park2018multimodal} integrates LSTM with variational auto-encoder that fuses signals and reconstructs expected distribution. For encoding, it projects multivariate observations and their temporal dependencies at each time step into a latent space using an LSTM-based encoder. For decoding, it estimates the expected distribution of multivariate inputs from the latent representation. OmniAnomaly~\cite{su2019robust} argues that deterministic methods may be misled by unpredictable instances and proposes a stochastic model for multivariate time-series anomaly detection. It captures the normal patterns behind data by learning robust representations of multivariate time-series with stochastic variable connection and planar normalizing flow. Their model considers patterns with low reconstruction probability as anomalies. 

As introduced above, both forecasting-based and reconstruction-based models have shown their superiority in some specific situations. The forecasting-based model is specialized for feature engineering of next timestamp prediction, and construction-based model is good at capturing the data distribution of entire time-series. In our paper, we demonstrate that they are complementary to each other empirically. Moreover, none of the existing solutions capture the correlations between multiple features explicitly, which is emphatically addressed in this paper to enhance the performance of multivariate time-series anomaly detection. 

%% file: tex/methodology.tex
\section{Methodology}
Multiple univariate time-series from the same entity forms a multivariate time-series. Multivariate time-series anomaly detection aims to detect anomalies at entity-level \cite{su2019robust}. The problem can be defined as follows. 

\newtheorem{defi}{Problem Definition}
\begin{defi}
\label{def}
	An input of multivariate time-series anomaly detection is denoted by $x \in R^{n \times k}$, where $n$ is the maximum length of timestamps, and $k$ is the number of features in the input. For a long time-series, we generate fixed-length inputs by a sliding window of length $n$. The task of multivariate time-series anomaly detection is to produce an output vector $y \in R^{n}$, where $y_i \in \{0, 1\}$ denotes whether the $i^{th}$ timestamp is an anomaly.
\end{defi}

We address this problem by modeling the inter-feature correlations and temporal dependencies with two graph attention networks in parallel, followed by a Gated Recurrent Unit (GRU) network to capture long-term dependencies in the sequential data. We also leverage the power of both forecasting-based and reconstruction-based models by optimizing a joint objective function. The following of this section is organized as follows. First, we will have a brief overview of our network in section \ref{sec:overview}. Then, the details of data prepossessing, graph attention layers, and joint optimization will be presented in Section \ref{sec:preprocessing}, \ref{sec:graph_attention}, and \ref{sec:joint_optimization} respectively. As last, the procedure of model inference is described in Section \ref{sec:341}. Table \ref{notations} summarizes the notations used in our model.

\begin{figure*}[htbp]
	\centering
	\includegraphics[width=\textwidth]{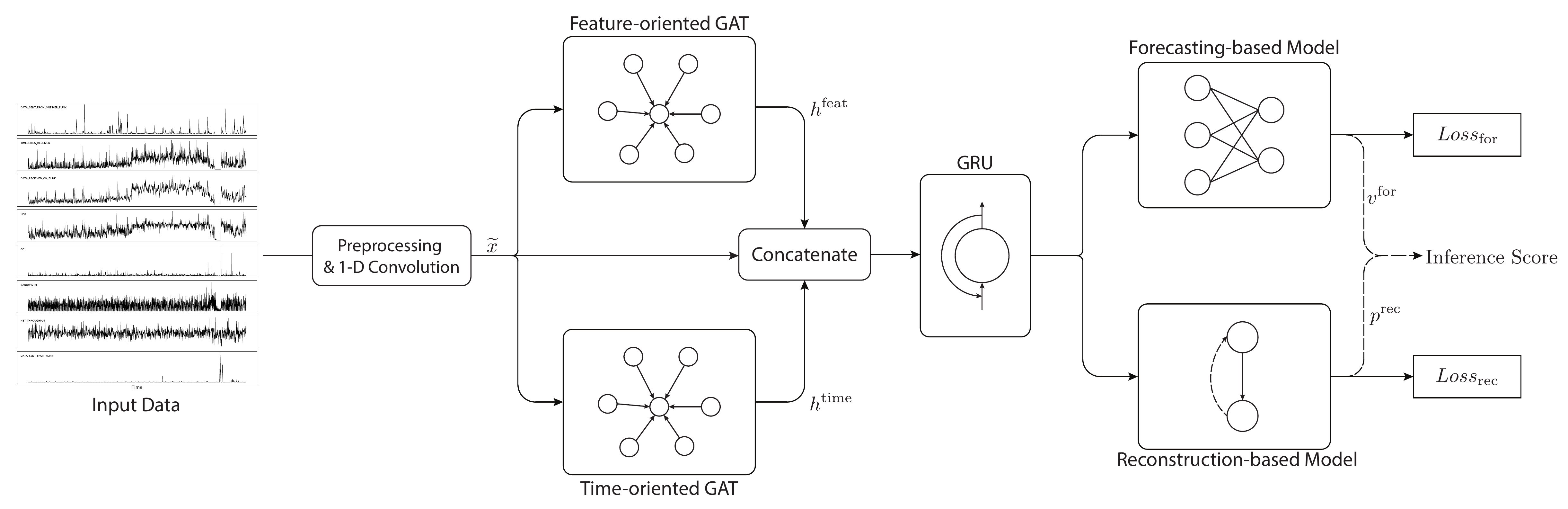}
	\caption{The architecture of MTAD-GAT for multivariate time-series anomaly detection}
	\label{network}
\end{figure*}

\begin{table}
  \caption{Notations}
  \label{notations}
 \begin{tabular}{p{0.5cm}p{7.0cm}}
    \toprule
%     Notation & \space \space Description \\
%    \midrule
  $x$ & an instance of multivariate time-series input \\
  $n$ & the length of $x$ in a pre-defined sliding window \\
  $k$ & the number of features (variates) in $x$\\ 
  $\widetilde{x}$ & input after preprocessing \\
  %$p,q$& sample number of training set and testing set\\
  %$S$ & step in sampling \\
  %$\eta$ & threshold of Spectral Residual (SR) for data cleaning\\
  %$s$ & kernal size of 1-D convolution \\ 
  $v_i$ & input node representation for a GAT layer\\
  $h_i$ & output node representation for a GAT layer\\
  $\alpha_{ij}$ & attention score of node $j$ to node $i$ in a GAT layer\\
  $d_1$ & hidden dimension of a GRU layer \\
  $d_2$ & hidden dimension of fully-connected layers in the forecasting-based model \\
  $d_3$ & latent space dimension of the VAE model \\
 % $\beta$ & hyper-parameter to balance the joint optimization target \\  
  $\gamma$ & hyper-parameter to combine multiple inference scores \\
 \bottomrule
\end{tabular}
\end{table}

\subsection{Overview}
\label{sec:overview}
%\begin{figure}
%	\centering
%	\includegraphics[height=8.0cm,width=7cm]{figures/GAT_fea.png}
%	\caption{Attention mechanism}
%	\label{attention}
%\end{figure}

The overall network architecture of MTAD-GAT is shown in Figure \ref{network}, which is composed of the following modules in order: 
\begin{enumerate}
    \item We apply a 1-D convolution with kernel size 7 at the first layer to extract high-level features of each time-series input. As demonstrated in previous work~\cite{dos2014deep}, convolution operations are good at local feature engineering within a sliding window.
    \item The outputs of 1-D convolution layer are processed by two parallel graph attention (GAT)~\cite{velivckovic2017graph} layers, which underline the relationships between multiple features and timestamps. 
    \item We concatenate the output representations from the 1-D convolution layer and two GAT layers, feed them into a Gated Recurrent Unit (GRU)~\cite{chung2014empirical} layer with $d_1$ hidden dimension. This layer is used for capturing sequential patterns in time-series.
    \item The outputs of the GRU layer are fed into a forecasting-based model and a reconstruction based model in parallel to obtain the final result. We implement the forecasting-based model as a fully-connected network, and adopt VAE~\cite{kingma2013auto} for the reconstruction-based model.
\end{enumerate}

\subsection{Data Preprocessing}
\label{sec:preprocessing}
To improve the robustness of our model, we perform data normalization and cleaning for each individual time-series. Data normalization is applied on both training and testing set, while cleaning is only applied on the training set. 

%Irregular and abnormal data in each individual stream may corrupt the inter-dimensional correlations and the temporal dependencies, while an unsupervised model is especially sensitive to them.  

\subsubsection{Data normalization}
We normalize the time-series with the maximum and minimum values from the training data:
\begin{align}
   \widetilde{x} =\frac{x-\min(X_\text{train})}{\max(X_\text{train})-\min(X_\text{train})}\label{eq:rel1}  
\end{align}where $\max(X_\text{train})$ and $\min(X_\text{train})$ are the maximum value and the minimum value of the training set respectively.
\subsubsection{Data cleaning}
Prediction-based and reconstruction-based models are sensitive to irregular and abnormal instances in the training data. To alleviate this problem, we employ a state-of-the-art univariate anomaly detection method, Spectral Residual (SR) \cite{ren2019time}, to detect anomaly timestamps in each individual time-series in the training data. Following \cite{ren2019time}, we set the threshold as 3 to generate anomaly detection results. Those detected anomaly timestamps will be replaced with normal values around that timestamp. Note that SR is light-weighted and adds little overhead to the entire model. 
%Irregular and abnormal data in each individual stream may corrupt the inter-dimensional correlations and the temporal dependencies, while an unsupervised model is especially sensitive to them.  
%Spectral Residual (SR) \cite{ren2019time} is an efficient and effective model for unsupervised single-variant time-series anomaly detection, which consists of three steps: (1) a Fourier transform to get the log amplitude spectrum; (2) calculation of the spectral residual; and (3) an inverse Fourier transform to get back to the time domain. After inverse Fourier transformation, the abnormal points can be largely amplified and we adopt a pre-defined threshed $\delta$ to filter out these points. SR is lightweight and completely unsupervised, adding little overhead to the entire model.
\subsection{Graph Attention}
\label{sec:graph_attention}
%\begin{figure}
%	\centering
%	\includegraphics[height=8.0cm,width=7cm]{figures/GAT_fea.png}
%	\caption{Attention mechanism}
%	\label{attention}
%\end{figure}
Here we introduce the graph attention (GAT) layers in detail, which are the core designs of MTAD-GAT Net. 
%We exploit the Graph Attention (GAT) layer to encode the multivariate time-series. Its structure is shown in Figure \ref{attention}.
%\subsubsection{Graph Attention Network}
A GAT layer is able to model the relationships between nodes in an arbitrary graph. Generally, given a graph with $n$ nodes, i.e., $\{v_1, v_2, \cdots, v_n\}$, where $v_i$ is the feature vector of each node, a GAT layer computes the output representation for each node as follows: 
\begin{equation}
h_i=\sigma(\sum_{j=1}^L \alpha_{ij}v_j),
\end{equation}
where $h_i$ denotes the output representation of node $i$, which has the same shape with input $v_i$; 
$\sigma$ represents the sigmoid activation function; $\alpha_{ij}$ is the attention score which measures the contribution of node $j$ to node $i$, where $j$ is one of the adjacent nodes for node $i$; $L$ denotes the number of adjacent nodes for node $i$. 

% {\color{red}
The attention score $\alpha_{ij}$ can be computed by the following equations: 
\begin{align}
	e_{ij}&=\text{LeakyReLU}(w^\top\cdot (v_i\oplus v_j)) \\
	\alpha_{ij}&=\frac{\text{exp}(e_{ij})}{\sum_{l=1}^{L}\text{exp}(e_{il})}.
	\label{eq:attention}
\end{align}

Here $\oplus$ represents concatenation of two node representations, $w\in R^{2m}$ is a column vector of learnable parameters where $m$ is the dimension of the feature vector of each node, and LeakyReLU is a nonlinear activation function~\cite{xu2015empirical}.
% }
In the multivariate time-series anomaly detection scenario, we exploit two types of graph attention layers, namely \textit{feature-oriented graph attention} and \textit{time-oriented graph attention}. 
\subsubsection{Feature-oriented graph attention layer}
\begin{figure}
	\centering
	\includegraphics[width=0.4\textwidth]{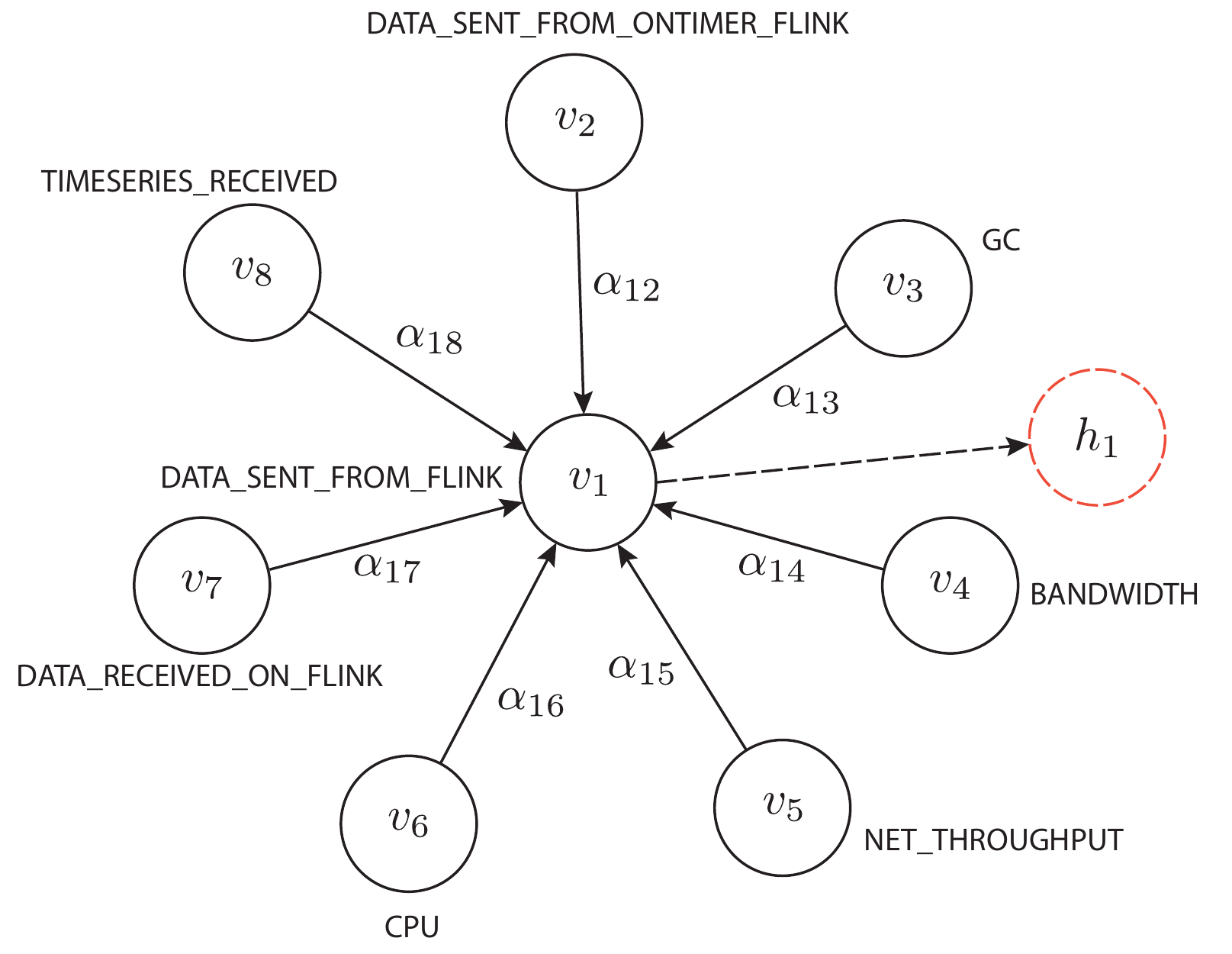}
	\caption{Feature-oriented graph attention layer. Dashed circle is the final output.}
	\label{graph_attention}
\end{figure}

On one hand, we need to detect multivariate correlations without any prior. Therefore, we treat the multivariate time-series as a complete graph, where each node represents a certain feature, and each edge denotes the relationship between two corresponding features. In this way, the relationships between adjacent nodes can be carefully captured through graph attention operations. Specifically, each node $x_i$ is represented by a sequential vector $x_i = \{x_{i,t} | t \in [0, n)\}$ and there are totally $k$ nodes, where $n$ is the total number of timestamps and $k$ is the total number of multivariate features.
% We assume no prior topology on the graph, so we consider the graph as a fully connected graph where each node has $K-1$ adjacent nodes. 
The layer is illustrated in Figure \ref{graph_attention}.
\subsubsection{Time-oriented graph attention layer}
On the other hand, we leverage the power of graph attention network to capture temporal dependencies in time-series. We consider all the timestamps within a sliding window as a complete graph. Concretely, a node $x_t$ represents the feature vector at timestamp $t$, and its adjacent nodes include all other timestamps in the current sliding window. This is much like a Transformer~\cite{vaswani2017attention}, where all words in a sequence are modeled by a fully-connected self-attention operation. 
%We compute the embed vector by $v_i=Wx_i$, where $W\in R^{M\times W}$ is the weight matrix.
%The feature oriented layer models the causal relationship between different features, thus each node stands for the corresponding time-series of a single feature dimension. Specifically, each node is represented by a feature vector $x_i\in R^W$ and there are in total $K$ nodes, where $K$ is the number of features. We assume no prior topology on the graph, so we consider the graph as a fully connected graph. Thus, each node has $K-1$ adjacent nodes. We compute the embed vector by $v_i=Wx_i$, where $W\in R^{M\times W}$ is the weight matrix.
%\subsubsection{Time-oriented Layer}
% {\color{red}
The output of the feature-oriented graph attention layer is a matrix with shape $k\times n$, where each row is an $n$ dimensional vector representing the output for each node and there are in total $k$ nodes. Similarly, the output of the time-oriented graph attention layer has a shape of $n\times k$. We concatenate the outputs from the feature-oriented graph attention layer and the time-oriented graph attention layer as well as the preprocessed $\tilde{x}$ to a matrix with shape $n\times 3k$, where each row represents a $3k$ dimensional feature vector for each timestamp, to fuse the information from different sources.
% }
\subsection{Joint Optimization}
\label{sec:joint_optimization}
As mentioned previously, the forecasting and the reconstruction models have distinct advantages respectively which are complementary with each other. Our model includes a \textit{forecasting-based} model to predict the value at next timestamp and a \textit{reconstruction-based} model to capture the data distribution of entire time-series. During the training process, the parameters from both models are updated simultaneously. 
%The loss function is the sum of two optimization targets, which can be defined as:
%\begin{equation}
%\label{eq:rel5} 
%Loss = Loss_{for}+\beta\times Loss_{rec}. 
%\end{equation}
The loss function is defined as the sum of two optimization targets, i.e., $Loss = Loss_{for} + Loss_{rec}$, where $Loss_{for}$ denotes the loss function of forecasting-based model and $Loss_{rec}$ denotes the loss function of reconstruction-based model. 
%$\beta$ is a hyper parameter to balance the influence of forecasting-based and reconstruction-based models, which is chosen on the validation set. We will also provide a sensitivity study in the analysis section.

\subsubsection{Forecasting-based model}
The forecasting-based model predicts the value at next timestamp. We stack three fully-connected layers with hidden dimensions $d_2$ after the GRU layer as the forecasting-based model. The loss function can be formulated as Root Mean Square Error (RMSE): 
\begin{align}
& Loss_{for}=\sqrt{\sum_{i=1}^{k}(x_{n,i}-\hat{x}_{n,i})^2}. \label{eq:rel6} 
\end{align}
where $x_n$ denotes the next timestamp for the current input $x = (x_0, x_1, ..., x_{n-1})$; $x_{n,i}$ represents the value for the $i^{th}$ feature in $x_n$; and $\hat{x}_{n,i}$ is the value predicted by the forecasting-based model. 
\subsubsection{Reconstruction-based model}
The reconstruction-based model aims to learn a marginal distribution of data over a latent representation $z$. We employ Variational Auto-Encoder (VAE)~\cite{kingma2013auto}, which provides a \textit{probabilistic} manner for describing an observation in the latent space. By treating the values of time-series as variables, the VAE model is able to capture the data distribution of entire time-series. Given an input $x$, it is supposed to be reconstructed from a conditional distribution $p_\theta(x|z)$, where $z \in R^{d_3}$ is the vector representation in a latent space. 
The optimization target is to find the best model parameters that reconstruct $x$ with the most close data distribution. The true posterior density can be given by:
\begin{equation}
p_\theta(z|x)=p_\theta(x|z)p_{\theta}(z)/p_{\theta}(x)
\end{equation}
where the marginal density is formulated as
\begin{equation}
p_\theta(x)=\int{p_\theta(z)p_\theta(x|z)dz}
\end{equation}
It is intractable to calculate the above equation, so we need to introduce a recognition model $q_\phi(z|x)$ to approximate the posterior distribution.
Given the recognition model (encoder) $q_\phi(z|x)$ and the generative model (decoder) $p_\theta(\hat{x}|z)$, the reconstruction-based loss function can be computed as follows:
%In our solution, the aforementioned network is treated as recognition model (data encoder), which generates a latent representation vector $z$. 
%Let $\phi$ and $\theta$ be the parameters of the recognition and generative models. 
\begin{equation}
\small{Loss_{rec} = -\mathrm{E}_{q_{\phi}(z|x)}[log{p_{\theta}(x|z)}] + D_{KL}(q_{\phi}(z|x)||p_{\theta}(z))}
\end{equation}
where the first term is the expected negative log-likelihood of the given input. The second term is the Kullback-Leibler divergence between the encoder’s distribution $q_{\phi}(z|x)$ and $p_{\theta}(z)$, which can be viewed as a regularizer. Theoretically, the negation of this loss function is a practical estimator of the lower bound for the intractable log likelihood, $\log{p_\theta(x)}$, so we can differentiate and optimize this loss function instead.
%And by yielding generic Stochastic Gradient Variational Bayes (SGVB) estimator and Monte Carlo integration the loss function can be rewritten as:
%\begin{align}
%Loss_{pro}  &=\mathrm{E}_{q_{\phi}(z|x^{(i)\ })}[log(p_\theta(z_t))-log(q_\phi(z_t|x_t))+log\ p_\theta(x^{(i)}|z)].  \nonumber \\6
%&\approx \frac{1}{L}\sum_{l=1}^{L}[log(p_\theta(z_t^{(l)})-log(q_\phi(z_t^{(l)}|x_t))+log(p_\theta(x_t|z_t^{(l)}))]. \label{eq:rel8}
%\end{align}
\subsection{Model Inference}
\label{sec:341}
Corresponding to the joint optimization target, we also have two inference results for each timestamp. One is prediction value $\{\hat{x}_i|i = 1,2,...,k\}$ calculated by the forecasting-based model, and the other is reconstruction probability $\{p_i|i = 1,2,..,k\}$ obtained from the reconstruction-based model. The final inference score balances their benefits to maximize the overall effectiveness of anomaly detection. We calculate an inference score $s_i$ for each feature and take the summation of all features as the final inference score. We identify a timestamp as an anomaly if its corresponding inference score is larger than a threshold. We use Peak Over Threshold (POT)~\cite{siffer2017anomaly} to choose the threshold automatically. Specifically, the inference score can be calculated by: 
%When a put time-series $X=\{x^{(i)}\}_{i=1}^{N}$ into our model, we will get a prediction loss RMSE  $Pre$ and estimate the reconstruction probability $Pro$ for $x^{N+1}$. 
%To let the model to learn the randomness of variables and to have the stability of the prediction model, we used the reconstruction probability and the weight of the prediction loss as the score of the anomaly detection.
\begin{align}
score = \sum_{i=1}^k s_i = \sum_{i=1}^k{\frac{(\hat{x}_{i}-x_{i})^2 + \gamma \times  (1 - p_{i})}{1 + \gamma}}
\label{eq:rel9}
\end{align}
where $(\hat{x}_i-x_i)^2$ is the squared error between the forecasting value $\hat{x}_i$ and the actual value $x_i$, indicating how much the actual value of feature $i$ is deviated from prediction; $(1 - p_i)$ is the probability of encountering an abnormal value for feature $i$ according to the reconstruction model; $k$ is the total number of features; and $\gamma$ is a hyper-parameter to combine the forecasting-based error and the reconstruction-based probability. $\gamma$ is chosen by grid search on the validation set, and a sensitivity study will be provided in the analysis section.

%% file: tex/experiments.tex
\section{Experiments}
\subsection{Datasets and Metrics}
\paragraph{Datasets.} We use three datasets to verify the effectiveness of our model, namely SMAP (Soil Moisture Active Passive satellite), MSL (Mars Science Laboratory rover) and TSA (Time Series Anomaly detection system). SMAP and MSL are spacecraft datasets collected by NASA \cite{o2010nasa}. TSA is a dataset collected from our own time-series anomaly detection system that processes time-series by Flink\footnote{https://flink.apache.org/}. We gathered two months metrics of services and hardwares from the Flink cluster of the system, and an example is shown in Figure \ref{fig:intro}. Anomaly labels used for evaluation in TSA dataset have been labeled based on incident reports from the system. The statistics of these three datasets are shown in Table \ref{tab:dataset}. 
%HWS is a 1-minute-long dataset from our own server. Its features contain server indicators like $CPU$,$Memory$, and the anomalies are labeled by domain experts.For experiment our dataset is divided into training set and testing set  according to 1:1 ratio.

% {\color{red}
\paragraph{Metrics.} We use precision, recall and F1-score to indicate the performance of our model.  Experience about AUC scores are also conducted. However we found it is hard to measure the performance by AUC for most state-of-the-art methods including ours have more than 0.97 AUC scores. In practice, anomalous observations usually form contiguous segment since they occur in a continuous manner. Following the evaluation strategy in \cite{su2019robust}, we treat the whole segment as correct if any observation in this segment is detected as anomaly correctly. 
% }

\begin{table}
	\caption{Dataset Statistics}
	\label{tab:dataset}
	\begin{tabular}{p{3.2cm}p{1.2cm}p{1.2cm}p{1.2cm}}
		\toprule
		\qquad &SMAP & MSL& TSA\\
		\midrule
		Number of sequences& 25& 55&18 \\
		Training set size& 135183 & 58317 &39312\\
		Testing set size& 427617& 73729&51408\\
		Anomaly Rate(\%)&13.13&10.27&10.58 \\
		\bottomrule
	\end{tabular}
\end{table}

\subsection{Setup}
We compare MTAD-GAT with state-of-the-art models for multivariate time-series anomaly detection, including OmniAnomaly \cite{su2019robust}, LSTM-NDT \cite{hundman2018detecting}, KitNet \cite{mirsky2018kitsune}, DAGMM \cite{zong2018deep}, GAN-Li \cite{li2018anomaly}, MAD-GAN \cite{li2019mad} and LSTM-VAE \cite{park2018multimodal}. We use the same sliding window size $n=100$ for all models. In our method, we set $\gamma=0.8$ through a grid search on the validation set. The hidden dimension sizes of the GRU layer ($d_1$), the fully-connected layers ($d_2$), and the VAE model ($d_3$) are set as 300 empirically. We use the Adam optimizer to train our model for 100 epochs with an initial learning rate 0.001. 
%\textcolor{red}{weight decay? L2 regularization?}
We compare the performance of state-of-the-art models in Section \ref{sec:SOTA}. To better understand our model, we also examine the effectiveness of different components through analysis in Section \ref{sec:ablation}.

\label{sec:SOTA}
\begin{table*}[htbp]
	\centering
	\caption{Performance of our models and baselines. }
	\begin{tabular}{cccccccccc}
		\toprule
		\multirow{2}{*}{Method} & \multicolumn{3}{c}{SMAP} & \multicolumn{3}{c}{MSL} & \multicolumn{3}{c}{TSA} \\
		\cmidrule(r){2-4} \cmidrule(r){5-7} \cmidrule(r){8-10}
		&  Precision      & Recall   &  F1
		&  Precision      & Recall   &  F1
		&  Precision      & Recall   &  F1  \\
		\midrule
		\multicolumn{10}{c}{Reconstruction based models}\\
		\midrule
		OmniAnomaly         &0.7416                      &0.9776                 &0.8434                  & 0.8867           & 0.9117          &0.8989         &0.7028  &0.8039&0.7499
    \\
		
		KitNet &0.7725&0.8327&0.8014&0.6312&0.7936&0.7031&0.5579&0.8012&0.6577\\
		GAN-Li&0.6710&0.8706&0.7579&0.7102&0.8706&0.7823&0.5302& 0.7551&0.6229\\
		MAD-GAN  &0.8049    &0.8214         &0.8131      &0.8517    &0.8991        &0.8747        &0.5510    &0. 8284  &0.6620\\
		LSTM-VAE           &0.8551                         & 0.6366                   & 0.7298                 & 0.5257           & 0.9546          &0.6780         &  0.6970     & 0.7736         &     0.7333   \\
		\midrule
		\multicolumn{10}{c}{Forecasting based models}\\
		\midrule
		LSTM-NDT             & 0.8965                        &0.8846             & 0.8905                 & 0.5934         & 0.5374          &0.5640         &   0.5833       & 0.7232          &    0.6457   \\
		DAGMM             &0.5845                         & 0.9058               & 0.7105              & 0.5412         &0.9934        &0.7007       &   0.5351       &  0.8845       &0.6668         \\
		\hline
		\hline
		\textbf{MTAD-GAT}     &0.8906                    &0.9123                   &\textbf{0.9013}                  & 0.8754       & 0.9440          &\textbf{0.9084}        &0.6951

        & 0.9352
         &\textbf{0.7975}            \\
		\bottomrule
	\end{tabular}
	\label{scores}
\end{table*}
%Quantitative results are showed in Table \ref{scores} and \ref{efficiency}. Table \ref{scores} summarizes evaluation performances of state-of-the-art models, and Table \ref{efficiency} compares our model with the best baseline under different delay criteria.
%In Table \ref{efficiency}, we compare the \textcolor{red}{efficiency} of our model with the best baseline.

\subsection{Comparison with SOTAs}
As shown in Table \ref{scores}, MTAD-GAT shows excellent generalization capability and achieves the best F1 scores consistently on three datasets. Specifically, we achieve 1\%, 1\%, and 9\% improvement over the best state-of-the-art performance on SMAP, MSL and TSA datasets respectively. These performance lifts are significant as verified by a hypothesis testing. %when all of the experiments share the same training set and testing set.

The limitation of OmniAnomaly lies in not addressing the feature correlations explicitly in the model, which is essential to the success of multivariate time-series anomaly detection. As introduced earlier, our feature-oriented GAT layer is by design to tackle this problem. The superiority of this layer is verified in our experiments, as our model outperforms OmniAnomaly significantly and consistently on all three datasets. 

The temporal information is also crucial for multivariate time-series anomaly detection. The performance of DAGMM is not ideal, because it dose not take the temporal information into consideration. In our model, GRU is used to capture long-term temporal dependencies, and a time-oriented GAT layer is applied to calculate attention scores between correlated timestamps. These designs are helpful to achieve a much better performance than DAGMM, and we also conduct additional experiments (shown in Section \ref{sec:ablation}) to compare different design variations. 

Forecasting-based methods, such as LSTM-NDT, have good performances on SMAP, but perform poorly on MSL and TSA datasets. They are sensitive to different scenarios because it cannot model unpredictable cases. On the other hand, reconstruction-based methods (for example, OmniAnomaly) achieve much better results on the MSL and TSA datasets. This implies that both \textit{forecasting-based} and \textit{reconstruction-based} models have their own advantages, and the joint optimization strategy proposed in this paper is beneficial to the final performance.

\subsection{Evaluation with Different Delays}

In practice, anomalies usually occur in a continuous segment and we require a model to detect them as soon as possible to take quick actions. Therefore, we compare our model with the current best baseline, OmniAnomaly under different delay metrics. We follow the evaluation protocol described in \cite{ren2019time}, that is, treating the whole segment as true positives, if and only if there is an anomaly point detected correctly and its timestamp is at most $\delta$ steps after the first anomaly of the segment. 

\begin{figure}[htbp]
	\centering
    \includegraphics[width=0.5\textwidth]{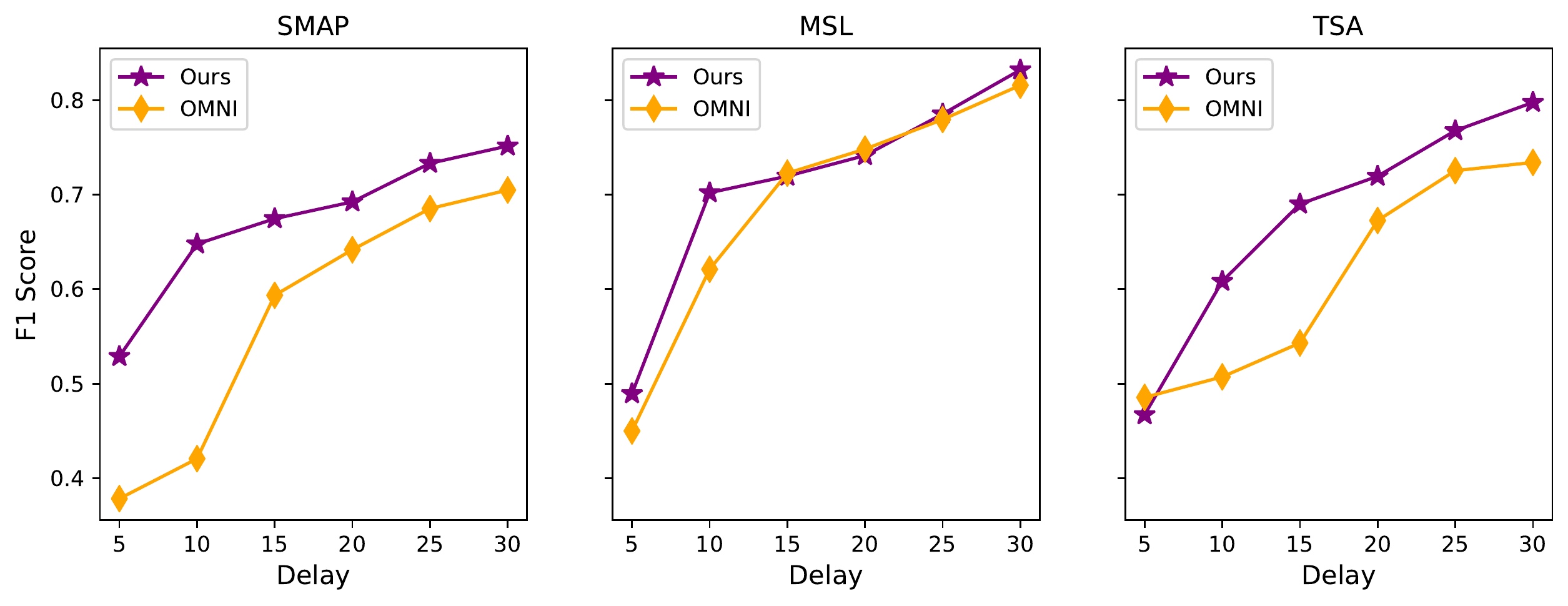}
	\caption{Comparison with different delay constraints}
	\label{fig:delay}
\end{figure}

Figure \ref{fig:delay} compares the F1-scores of our model and OmniAnomaly models for different delay metrics on three datasets. Notice that the F1 score becomes larger as the delay $\delta$ increases, and when $\delta$ is large enough, the number will match that reported in Table \ref{scores}. Overall, Our model achieves better performance consistently, especially when the acceptable delay is small. When $\delta=10$, the relative enhancements on these datasets are 53.98\%, 13.04\%, and 19.93\%. We can also observe a performance boost when the delay increases from 5 to 10 on all the three datasets. Therefore, the proposed anomaly detection model based on graph attention network is able to alert for potential incidents on time in real scenarios without excessive losses.

% \begin{table}
% 	\caption{Comparison on different delay constraints}
% 	\label{tab:commands}
% 	\begin{tabular}{cllccc}
% 		%     \toprule
% 		\hline
% 		Model&\multicolumn{2}{c}{Delay}&SMAP&MSL&HWS\\
% 		\hline
% 		\multirow{6}*{OMNI}
% 		&\multicolumn{2}{c}{5}&0.3784&0.4500&0.4855\\
% 		%      \cline{3-6}
% 		&\multicolumn{2}{c}{10}&0.4207&0.6210&0.5072\\
% 		%      \cline{3-6}
% 		&\multicolumn{2}{c}{15}&0.5935&0.7226&0.5431\\
%       	&\multicolumn{2}{c}{20}&0.6417&0.7481&0.6726\\
% 		&\multicolumn{2}{c}{25}&0.6852&0.7794&0.7253\\
% 		&\multicolumn{2}{c}{30}&0.7048&0.8156&0.7340\\
% 		\hline
% 		\multirow{6}*{GARN}
% 		&\multicolumn{2}{c}{5}&0.5288&0.4894&0.5111\\
% 		%      \cline{3-6}
% 		&\multicolumn{2}{c}{10}&0.6478&0.7020&0.5694\\
% 		%      \cline{3-6}
% 		&\multicolumn{2}{c}{15}&0.6745&0.7195&0.6445\\
%         &\multicolumn{2}{c}{20}&0.6923&0.7413&0.7159\\
% 		&\multicolumn{2}{c}{25}&0.7331&0.7852&0.7622\\
% 		&\multicolumn{2}{c}{30}&0.7514&0.8319&0.7696\\
% 		\hline
% 	\end{tabular}
% 	\label{efficiency}
% \end{table}
% \input{tex/qualitative_results}

\section{Analyses}
\label{sec:ablation}
In this section, we analyze the effectiveness of graph attention and joint optimization through comprehensive experiments (summarized in Table \ref{tab:ablation}), which will be discussed in detail with deeper insights. Then, we perform sensitivity study on the parameter $\gamma$ and present experimental study on anomaly diagnosis. 
%We conduct additional experiments to verify the superiority of these designs (summarized in Table \ref{tab:ablation}), 

\begin{table}
	\caption{Quantitative results for analyses. F1 scores are reported.}
	\label{tab:ablation}
	\begin{tabular}{p{3.5cm}p{1.2cm}p{1.2cm}p{1.2cm}}
		\toprule
		Model & SMAP & MSL & TSA \\
		\midrule
		\textbf{MTAD-GAT} & \textbf{0.9013} & \textbf{0.9084} & \textbf{0.7975} \\ \midrule
		w/o feature & 0.8783 & 0.8851 & 0.7474 \\
		w/o time & 0.8832 & 0.8897 & 0.7582 \\
		w/o prediction & 0.8731 & 0.8857 & 0.7380 \\
		w/o reconstruction & 0.8352 & 0.8058 & 0.7278 \\
%		w/o preprocessing& 0.8972 & 0.9011 & 0.7680 \\
%		w/o 1D-conv& 0.9002& 0.8983 & 0.7805 \\
%		\midrule
%		\textcolor{red}{manual threshold} & \textbf{0.9077} & \textbf{0.9201} & \textbf{0.7937} \\
		\bottomrule
		% &GARNw/oFeature& 0.8783 &0.8851 &0.7529\\
		%   & GARNw/oTime& 0.8832& 0.8897&0.7634\\
		%  &GARNw/oPred&0.8731&0.8857&0.7610\\
		%  & GARNw/oProb&0.8352&0.8058&0.7453
	\end{tabular}
\end{table}

\subsection{Effectiveness of Graph Attention}
We examine the influence of two graph attention layers in our model by disabling the feature-oriented GAT layer (denoted as \textit{w/o feature}) and the time-oriented GAT layer (denoted as \textit{w/o time}) one at a time. We adjust the number of parameters of both models to remove the impact of model complexity. From Table \ref{tab:ablation}, we find that \textit{w/o feature} causes 3.2\% decline in average F1 score, while \textit{w/o time} causes 2.5\% decline on it. Specifically, on TSA dataset, F1 score of \textit{w/o feature} drops by 4.8\% relatively compared to the full implementation. In fact, features like \textit{\footnotesize{CPU}} and \textit{\footnotesize{MEMORY}} are highly correlated on this dataset, and the feature-oriented GAT layer is useful to capture this correlation accurately for better anomaly detection. Moreover, the time-oriented GAT layer is also crucial to the final performance, although a GRU layer is already adopted for modeling the temporal dependencies. A potential explanation is that the time-oriented GAT layer can model the relationship between a pair of timestamps directly even if they are not adjacent. In this way, some long-term dependencies between timestamps can be modeled more explicitly. 
\begin{figure}[htbp]
	\centering
    \includegraphics[width=0.47\textwidth]{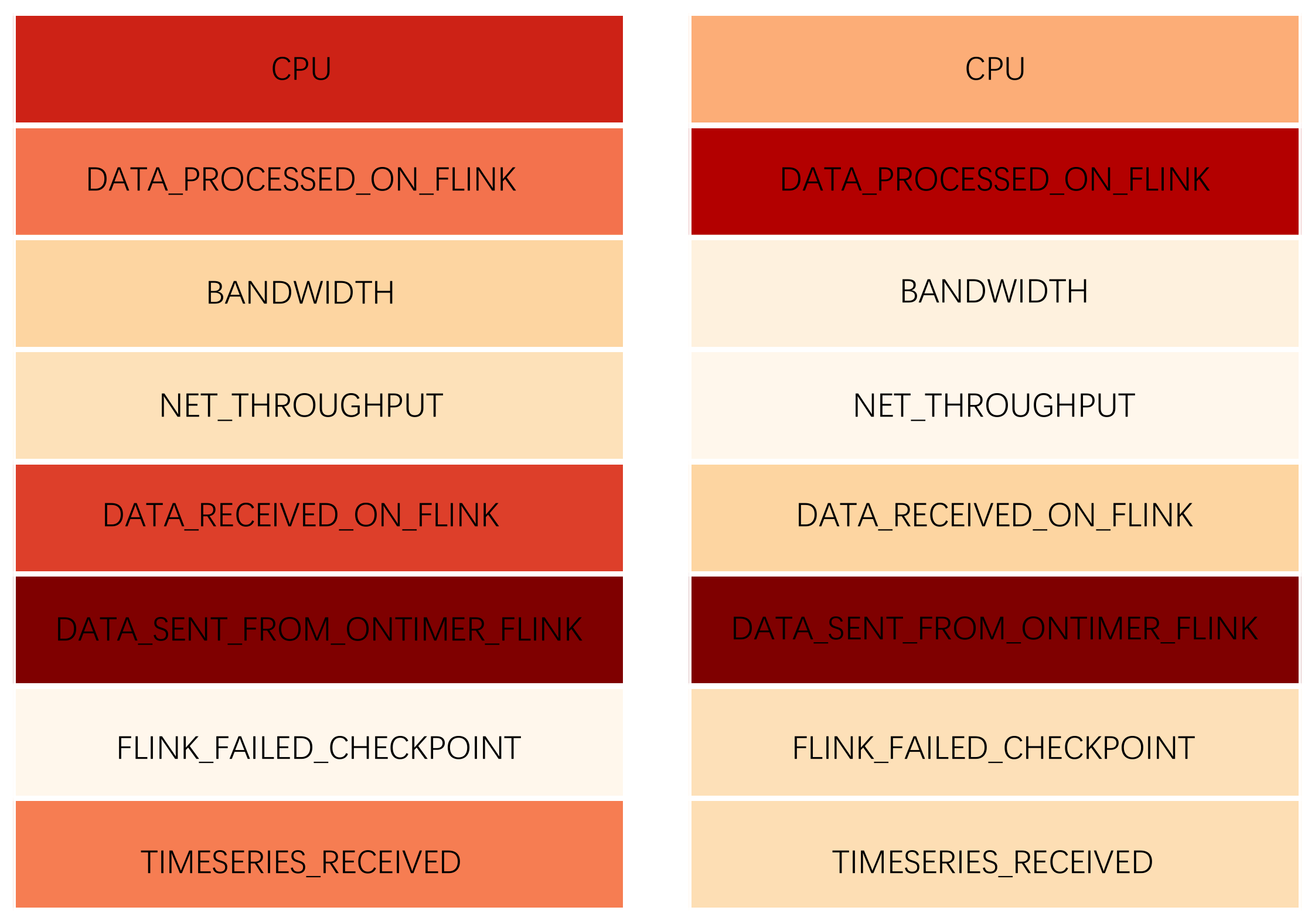}
	\caption{Illustration of attention scores for feature \textit{\footnotesize{DATA\_SENT\_FROM\_FLINK}} in the case of Figure 1. The left part visualizes the average attention scores at normal timestamps while the right part visualizes the attention scores when anomaly occurs. Darker color indicates higher attention scores.}
	\label{fig:corr}
\end{figure}

Here we leverage the example in Figure \ref{fig:intro} to explain why the feature-oriented GAT layer is helpful to the performance of anomaly detection. As the data source of TSA dataset is processed by Flink, a couple of features have been collected to reflect the running state of the system. \textit{\footnotesize{TIMESERIES\_RECEIVED}} indicates the number of time-series monitored in the system, which are sent to Flink for calculation and delivered to down-streaming components. \textit{\footnotesize{DATA\_RECEIVED\_ON\_FLINK}}, \textit{\footnotesize{DATA\_SENT\_FROM\_FLINK}}, and \textit{\footnotesize{CPU}} represent the data volume received on Flink, sent by Flink, and CPU utilization respectively. When the system works normally, other features should have strong positive correlation with \textit{\footnotesize{DATA\_SENT\_FROM\_FLINK}}. In Figure \ref{fig:corr}, we visualize the attention score $\alpha_{ij}$ calculated by the feature-oriented GAT layer based on Equation (\ref{eq:attention}). As illustrated in the left part of this figure, our model correctly learns the most relevant features to \textit{\footnotesize{DATA\_SENT\_FROM\_FLINK}} under normal circumstances. When anomaly occurs (corresponding to the red segment in Figure \ref{fig:intro}), the attention scores are visualized in the right part of Figure \ref{fig:corr}. We observe that the features \textit{\footnotesize{CPU}} and \textit{\footnotesize{DATA\_RECEIVED\_ON\_FLINK}} demonstrate much weaker correlations with \textit{\footnotesize{DATA\_SENT\_FROM\_FLINK}}. Actually, there is a traffic drop in the system, meanwhile, a garbage collection issue has been found on the Flink cluster. Thus, the job couldn't complete the checkpoint and keeps re-processing the last batch of the input stream. The continuously re-processed stream causes spike in the output metric, so \textit{\footnotesize{DATA\_SENT\_FROM\_FLINK}} has shown evident inconsistency.

\subsection{Effectiveness of Joint Optimization}
%  Table \ref{ablation:target} shows the experiments of dual optimization strategy:
In this section, we show the effectiveness of the joint optimization strategy by comparing F1 scores with controlled experiments. We compare our model with its simplified counterparts with single optimization target. Quantitative results in Table \ref{tab:ablation} show that the reconstruction-based variant (denoted as \textit{w/o prediction}) achieves better performance than the forecasting-based variant (denoted as \textit{w/o reconstruction}), but both of them degrade the original performance of the model significantly. 

\begin{figure}[t]
	\centering
    \includegraphics[width=0.48\textwidth]{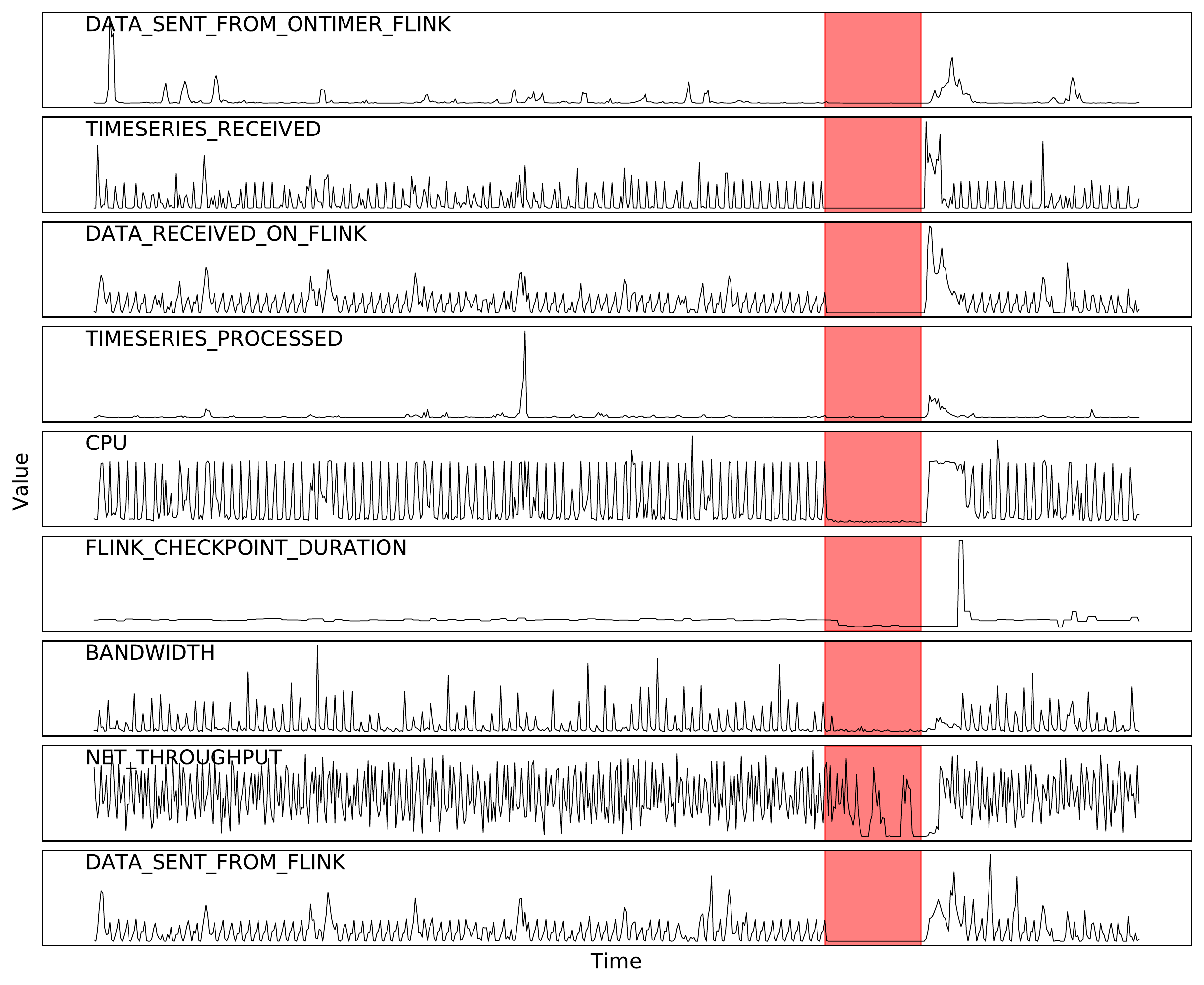}
	\caption{A failure case for reconstruction-based model}
	\label{fig:case}
\end{figure}

Forecasting-based model predicts the actual value of next timestamp in a deterministic manner, which is sensitive to randomness of time-series. 
%strategy emphasizes deviation between the predicted value and the actual value. In practice, time-series value could be random with various noise, thus it could be difficult to get a deterministic prediction value. 
On the other hand, reconstruction model alleviates this problem by learning a distribution of stochastic variables, which is more robust to perturbations and noises. However, there are still some cases that the reconstruction-based variant is not able to handle. For example, Figure \ref{fig:case} shows a periodic time-series containing an anomaly segment that the reconstruction-based variant does not find out. Generally, a reconstruction-based model is good at capturing global data distribution, but it may neglect sudden perturbations to disrupt periodicity in a time-series, especially when the values still conform to normal distribution. As shown in Figure \ref{fig:case}, the time-series has frequent periodicity in normal circumstances, so an anomaly should be detected when it is broken at the red segment. However, as the values still conform to normal distribution, the reconstruction-based variant fails to report the incident. Instead, this is successfully captured by the forecasting-based variant. Therefore, the joint optimization target is useful for achieving better anomaly detection results.
\subsection{Analysis of $\gamma$}
We perform additional experiments on analyzing the influence of $\gamma$ which balances the forecasting-based error and the reconstruction-based probability. We evaluate precision, recall, and F1 scores on the TSA dataset with different values of $\gamma$. Results are summarized in Table \ref{tab:gamma_vs_f1}. We notice that different settings of $\gamma$ achieve similar performance on all the three evaluation metrics. Specifically, when $\gamma=0.8$, we achieve the best performance. The result shows that our model is robust against $\gamma$. By setting $\gamma$ between 0.4 and 1.0, we can always achieve better results consistently than other state-of-the-art solutions listed in Table \ref{scores}. 

% 1.0	0.7849646246185829	0.9147389809315412	0.6874375991073002
% 0.8	0.7974810075969613	0.9352141294154424	0.6951092007434945
% 0.6	0.7851529844507606	0.9174742106908409	0.6861885557309019
% 0.4	0.7858330562021948	0.9233354173179118	0.6839759175639689

\begin{table}
    \centering
    \caption{Quantitative results for different $\gamma$ on TSA dataset}
    \label{tab:gamma_vs_f1}
    \begin{tabular}{p{1.8cm}p{1.5cm}p{1.9cm}p{1.9cm}}
        \toprule
         $\gamma$ & Precision & Recall & F1\\
         \midrule
         1.0 & 0.6874 & 0.9147 & 0.7849 \\
         0.8 & \textbf{0.6951} & \textbf{0.9352} & \textbf{0.7975} \\
         0.6 & 0.6861 & 0.9174 & 0.7851 \\
         0.4 & 0.6839 & 0.9233 & 0.7858 \\
         \bottomrule
    \end{tabular}
\end{table}
% \begin{figure}
%     \centering
%     \includegraphics[width=0.5\textwidth]{figures/gamma_vs_f1.pdf}
%     \caption{F1 scores with different $\gamma$ on the TSA dataset.}
%     \label{fig:gamma_vs_f1}
% \end{figure}

\subsection{Anomaly Diagnosis}
\label{sec:diagnosis}
Besides detecting anomalies in multivariate time-series, our method also provides useful insights for anomaly diagnosis. In real applications, those insights help people find the root cause of an incident and save the efforts to resolve it. For an instance of multivariate time-series, $x \in R^{n \times k}$, we define a collection of features, $\{x_i\} \in R^{n \times m} \subseteq x$ as the root causes, where $x_i \in R^{n}$ represents for the univariate time-series corresponding to a certain feature.

An algorithm of anomaly diagnosis sorts these features by the possibility of being the actual root cause for an incident. Then, it selects the top $m$ features to be the predicted root causes. In our method, we have a set of inference scores for each instance which can be calculated by Equation (\ref{eq:rel9}), that is, $S=\{s_1,s_2,...s_k\}$, where $k$ is the number of features and $s_i$ is the inference score for feature $i$. In our experiments, we select top 8 features with largest inference scores as root cause candidates.

We demonstrate diagnostic performance on our dataset TSA. We leverage our incident records to label root causes for each anomaly event. Along with the system developers and operators, we label more than 700 qualified instances in the test data which have clear root causes in the incident records. We use two metrics to evaluate the performance of anomaly diagnosis, HitRate@P\%~\cite{su2019robust} and NDCG~\cite{jarvelin2002cumulated}. HitRate is used to measure how many ground truths have been included in the top candidates. It can be calculated by $HitRate@P\%=\frac{Hit@\lfloor P\%\times|GT|\rfloor}{|GT|}$, where $|GT|$ is the number of ground truth for a single anomaly event. Normalized Discounted Cumulative Gain (NDCG) is a popular measure for relevance evaluation, and we adopt it here to quantify the ranking accuracy of root causes.

As shown in Table~\ref{tab:anomaly_interpretation}, our model demonstrates the capability of finding the top root cause features. Especially, 70\% true root cause has been captured in the diagnose results and the ranking performance of top 5 results also indicate our approach has a high probability to find the root cause at the top 5 candidates learned by the algorithm. In real AI-ops scenarios, this ability help us find the real causes and solve the problem as soon as possible.

The ability of anomaly diagnosis is much owing to the graph attention layer leveraged in the model. In real scenarios, a feature could be the actual root cause if its correlations with others have changed. For example in Figure \ref{fig:intro}, when there is a dip in the input stream, we expect a lower value in the output steam. Thus, features \textit{\footnotesize{DATA\_SENT\_FROM\_FLINK}} and \textit{\footnotesize{DATA\_RECEIVED\_FROM\_FLINK}} should keep consistent tendency. When there is an abnormal correlation between them, we can speculate that an incident occurs in the system, and these two features may be the potential root causes. The feature-oriented graph attention layer in our model captures the correlation between features properly, so this complicated circumstance can be well-handled.
\begin{table}
    \centering
    \caption{Quantitative results for anomaly diagnosis}
    \label{tab:anomaly_interpretation}
    \begin{tabular}{p{1.8cm}p{1.5cm}p{1.9cm}p{1.9cm}}
        \toprule
         Model & NDCG@5 & HitRate@100\% & HitRate@150\% \\
         \midrule
         \textbf{MTAD-GAT} & 0.8556 & 0.7428 & 0.8561 \\
         \bottomrule
    \end{tabular}
\end{table}

%% file: tex/qualitative_results.tex
\section{Case Study}
In this section, we provide another successful case and a failure case to analyze the advantages and deficiencies of our model. 
\begin{figure}[h!]
    \centering
    \subfigure[A true negative case that MTAD-GAT avoids false alarm]
    {
        \includegraphics[width=0.48\textwidth]{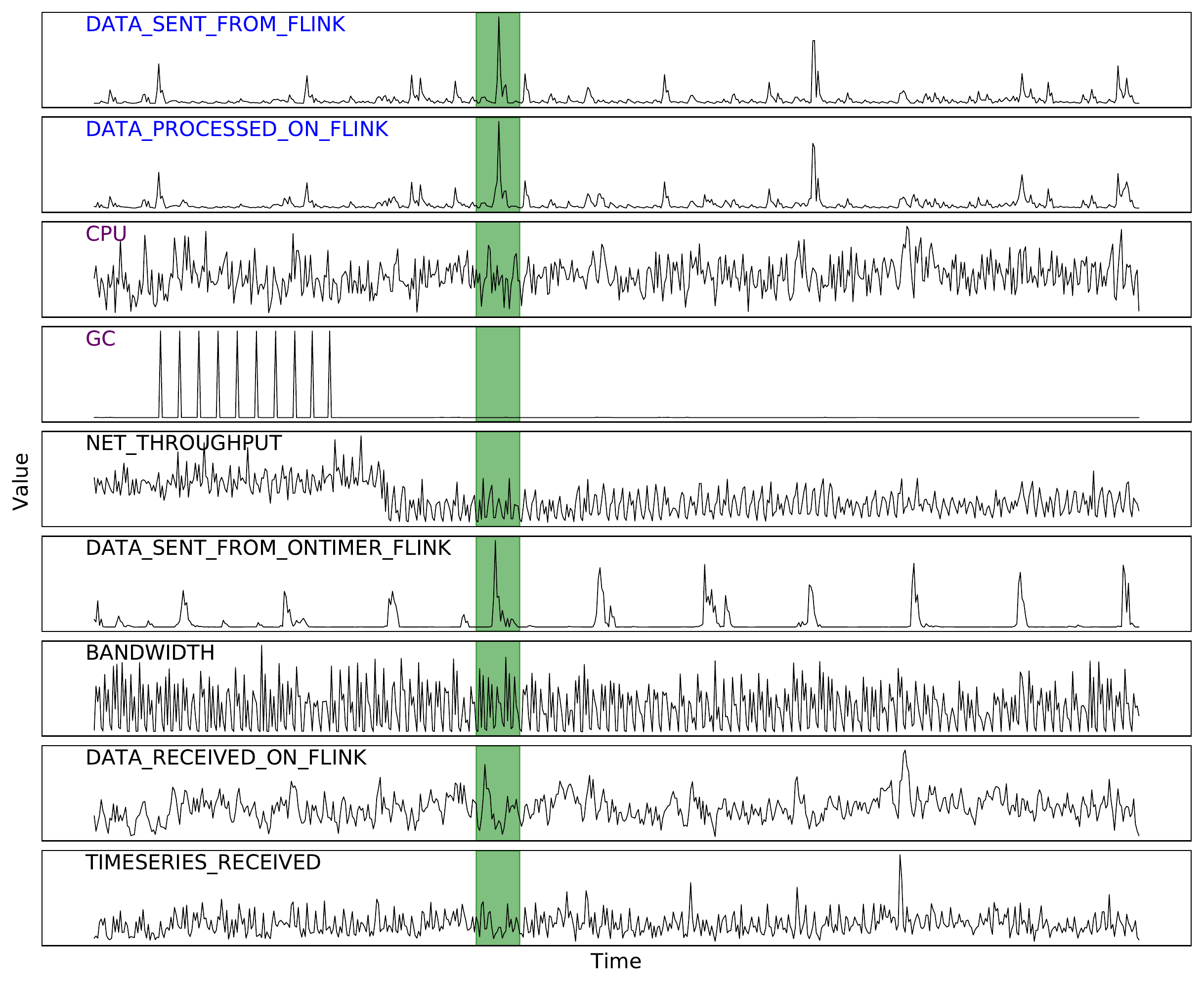}
        \label{fig:pos}
    }
    \subfigure[A false positive case that MTAD-GAT generates false alarm]
    {
        \includegraphics[width=0.48\textwidth]{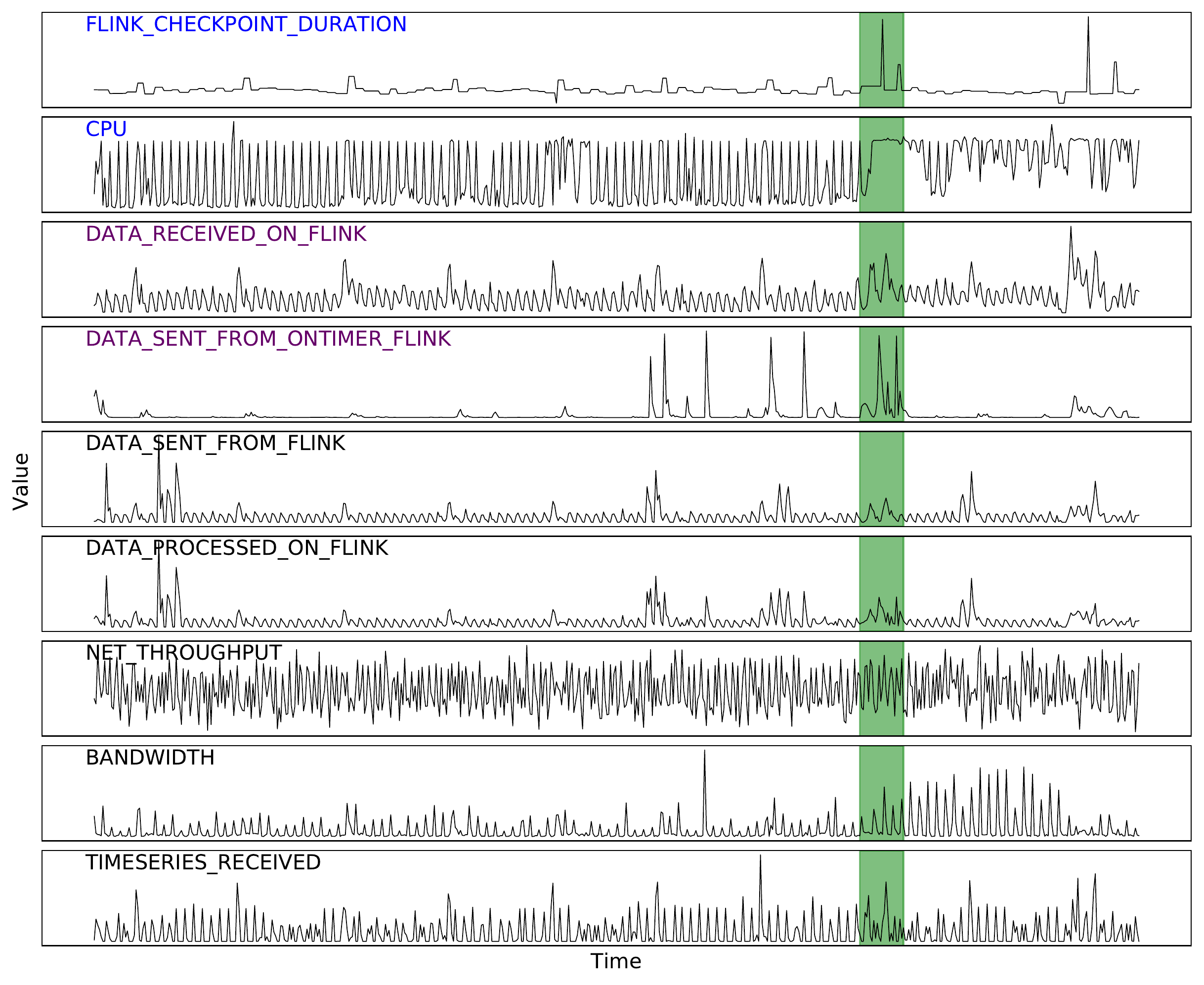}
        \label{fig:neg}
    }
    \caption{Case Study}
    \label{fig:cases}
\end{figure}
In Figure \ref{fig:pos}, we show a true negative case of our model. In the green segment, spikes show up in several time-series. If we detect each univariate time-series separately, we may alert for an abnormal incident. But considering feature relationships, we can find that the correlations between those features remain unchanged, although their volumes have greatly boosted. 
%This case can be detected as anomaly by models which only consider the value of multivariate time-series. In MAGA-Net, we leverage correlation between those series. By introducing correlation, our model can adapt to the consistent change of relative series. 
Actually, these spikes indicate that more data has been processed in the Flink job, and it is normal to have spikes in the corresponding time-series of \textit{\footnotesize{DATA\_SENT\_FROM\_FLINK}} and \textit{\footnotesize{DATA\_PROCESSED\_ON\_FLINK}}. 
Meanwhile, the values of \textit{\footnotesize{CPU}} and \textit{\footnotesize{GC}} keep stable, indicating that the system is in a healthy state and the increased traffic can be processed smoothly. To conclude, our approach is advantageous in dealing these cases and thus decreases the number of false alerts largely in our monitoring system. 

Next, we analyze a false positive case plotted in Figure \ref{fig:neg}. The green segment in the figure represents for a normal instance that is falsely detected as an anomaly by our model. Through detailed examination we find that our approach identifies abnormal events on \textit{\footnotesize{FLINK\_CHECKPOINT\_DURATION}} and \textit{\footnotesize{CPU}} because there are unusual spikes as shown in Figure \ref{fig:neg}. However, this is caused by an increase of input data volume and the related features such as \textit{\footnotesize{DATA\_RECEIVED\_ON\_FLINK}} and \textit{\footnotesize{DATA\_SENT\_FROM\_ONTIMER\_FLINK}} also demonstrate the same pattern. In this case, although a Flink job takes more time to complete the checkpoint, the traffic peak does not last for a long time, so it is not considered as an anomaly in the system. However, as this spike appears seldom in the history, the unsupervised anomaly detection algorithm naturally treats it as an abnormal case. We may need more domain knowledge or user feedbacks to solve this problem, which is left for future work.

%There are spikes at the time-series of \textit{\footnotesize{FLINK\_CHECKPOINT\_DURATION}} in both instances. However, related features, such as \textit{\footnotesize{DATA\_SENT\_FROM\_ONTIMER\_FLINK}}

%\textit{\footnotesize{TIMESERIES\_RECEIVED}} has spikes at the red segment. Correspondingly, related features, such as \textit{\footnotesize{DATA\_RECEIVED\_ON\_FLINK}} also show the same pattern. Therefore, there are no obvious changes on attention scores between those features. However, MAGA-Net captures anomalies on \textit{\footnotesize{CPU}} and \textit{\footnotesize{FLINK\_CHECKPOINT\_DURATION}}. Before that moment, \textit{\footnotesize{FLINK\_CHECKPOINT\_DURATION}} almost keeps in a stable value range. Thus significant and unusual spike makes the moment be detected as anomaly. Abnormal pattern on \textit{\footnotesize{CPU}} also helps confirm thus decision. In reality, although Flink job takes more time to complete the checkpoint, the traffic peak doesn't host for a long time, hence the system has idle time to tolerate the abnormal latency.

%% file: tex/conclusion.tex
\section{Conclusion}
In this paper, we propose a novel framework based on graph attention network for multivariate time-series anomaly detection. By learning feature-wise and temporal relationships of multivariate time-series and leveraging a joint optimization strategy, our method outperforms other state-of-the-art models on three datasets consistently. In addition, our model demonstrates good capability of anomaly diagnosis, which helps customers to find the actual root causes for an anomaly event. Extensive analysis provides more insights into the model and verifies the effectiveness of the proposed architecture. Future works may come from two aspects. First, our model has no prior knowledge on the correlations between features. Using user feedback or domain prior knowledge may benefit the performance. Second, current anomaly diagnosis is studied on relatively simple scenarios. We may utilize our model to investigate more complicated cases.
%In addition, we provide an innovative method for anomaly diagnosis by using the correlation scores between features to find the potential factors may result in the anomaly. 

%Thorough analysis has demonstrated that our model can capture complex relationships in multivariate time-series and can be applied in various scenarios.

%As multivariate anomaly detection has been an important topic in more field, we believe that our model with excellent performance can be applied in more scenarios.

%% file: maga.bbl
% Generated by IEEEtran.bst, version: 1.12 (2007/01/11)
\begin{thebibliography}{10}
\providecommand{\url}[1]{#1}
\csname url@samestyle\endcsname
\providecommand{\newblock}{\relax}
\providecommand{\bibinfo}[2]{#2}
\providecommand{\BIBentrySTDinterwordspacing}{\spaceskip=0pt\relax}
\providecommand{\BIBentryALTinterwordstretchfactor}{4}
\providecommand{\BIBentryALTinterwordspacing}{\spaceskip=\fontdimen2\font plus
\BIBentryALTinterwordstretchfactor\fontdimen3\font minus
  \fontdimen4\font\relax}
\providecommand{\BIBforeignlanguage}[2]{{%
\expandafter\ifx\csname l@#1\endcsname\relax
\typeout{** WARNING: IEEEtran.bst: No hyphenation pattern has been}%
\typeout{** loaded for the language `#1'. Using the pattern for}%
\typeout{** the default language instead.}%
\else
\language=\csname l@#1\endcsname
\fi
#2}}
\providecommand{\BIBdecl}{\relax}
\BIBdecl

\bibitem{ren2019time}
H.~Ren, B.~Xu, Y.~Wang, C.~Yi, C.~Huang, X.~Kou, T.~Xing, M.~Yang, J.~Tong, and
  Q.~Zhang, ``Time-series anomaly detection service at microsoft,'' in
  \emph{Proceedings of the 25th ACM SIGKDD International Conference on
  Knowledge Discovery \& Data Mining}, 2019, pp. 3009--3017.

\bibitem{hundman2018detecting}
K.~Hundman, V.~Constantinou, C.~Laporte, I.~Colwell, and T.~Soderstrom,
  ``Detecting spacecraft anomalies using lstms and nonparametric dynamic
  thresholding,'' in \emph{Proceedings of the 24th ACM SIGKDD International
  Conference on Knowledge Discovery \& Data Mining}, 2018, pp. 387--395.

\bibitem{malhotra2016lstm}
P.~Malhotra, A.~Ramakrishnan, G.~Anand, L.~Vig, P.~Agarwal, and G.~Shroff,
  ``Lstm-based encoder-decoder for multi-sensor anomaly detection,''
  \emph{arXiv preprint arXiv:1607.00148}, 2016.

\bibitem{su2019robust}
Y.~Su, Y.~Zhao, C.~Niu, R.~Liu, W.~Sun, and D.~Pei, ``Robust anomaly detection
  for multivariate time series through stochastic recurrent neural network,''
  in \emph{Proceedings of the 25th ACM SIGKDD International Conference on
  Knowledge Discovery \& Data Mining}, 2019, pp. 2828--2837.

\bibitem{velivckovic2017graph}
P.~Veli{\v{c}}kovi{\'c}, G.~Cucurull, A.~Casanova, A.~Romero, P.~Lio, and
  Y.~Bengio, ``Graph attention networks,'' \emph{arXiv preprint
  arXiv:1710.10903}, 2017.

\bibitem{wong2015rad}
J.~Wong, C.~Colburn, E.~Meeks, and S.~Vedaraman, ``Rad—outlier detection on
  big data,'' \emph{Web blog post. The Netflix Tech Blog. Netflix}, vol.~19,
  2015.

\bibitem{kejariwal2015introducing}
A.~Kejariwal, ``Introducing practical and robust anomaly detection in a time
  series,'' \emph{Twitter Engineering Blog. Web}, vol.~15, 2015.

\bibitem{malhotra2015long}
P.~Malhotra, L.~Vig, G.~Shroff, and P.~Agarwal, ``Long short term memory
  networks for anomaly detection in time series,'' in \emph{Proceedings}.\hskip
  1em plus 0.5em minus 0.4em\relax Presses universitaires de Louvain, 2015,
  p.~89.

\bibitem{zong2018deep}
B.~Zong, Q.~Song, M.~R. Min, W.~Cheng, C.~Lumezanu, D.~Cho, and H.~Chen, ``Deep
  autoencoding gaussian mixture model for unsupervised anomaly detection,'' in
  \emph{International Conference on Learning Representations}, 2018.

\bibitem{li2018anomaly}
D.~Li, D.~Chen, J.~Goh, and S.-K. Ng, ``Anomaly detection with generative
  adversarial networks for multivariate time series,'' \emph{arXiv preprint
  arXiv:1809.04758}, 2018.

\bibitem{li2019mad}
D.~Li, D.~Chen, L.~Shi, B.~Jin, J.~Goh, and S.-K. Ng, ``Mad-gan: Multivariate
  anomaly detection for time series data with generative adversarial
  networks,'' \emph{arXiv preprint arXiv:1901.04997}, 2019.

\bibitem{park2018multimodal}
D.~Park, Y.~Hoshi, and C.~C. Kemp, ``A multimodal anomaly detector for
  robot-assisted feeding using an lstm-based variational autoencoder,''
  \emph{IEEE Robotics and Automation Letters}, vol.~3, no.~3, pp. 1544--1551,
  2018.

\bibitem{mirsky2018kitsune}
Y.~Mirsky, T.~Doitshman, Y.~Elovici, and A.~Shabtai, ``Kitsune: an ensemble of
  autoencoders for online network intrusion detection,'' \emph{arXiv preprint
  arXiv:1802.09089}, 2018.

\bibitem{ding2018multivariate}
N.~Ding, H.~Gao, H.~Bu, H.~Ma, and H.~Si, ``Multivariate-time-series-driven
  real-time anomaly detection based on bayesian network,'' \emph{Sensors},
  vol.~18, no.~10, p. 3367, 2018.

\bibitem{gugulothu2018sparse}
N.~Gugulothu, P.~Malhotra, L.~Vig, and G.~Shroff, ``Sparse neural networks for
  anomaly detection in high-dimensional time series,'' in \emph{AI4IOT workshop
  in conjunction with ICML, IJCAI and ECAI}, 2018.

\bibitem{mousavi2015analyzing}
H.~Mousavi, S.~Mohammadi, A.~Perina, R.~Chellali, and V.~Murino, ``Analyzing
  tracklets for the detection of abnormal crowd behavior,'' in \emph{2015 IEEE
  Winter Conference on Applications of Computer Vision}.\hskip 1em plus 0.5em
  minus 0.4em\relax IEEE, 2015, pp. 148--155.

\bibitem{rosner1983percentage}
B.~Rosner, ``Percentage points for a generalized esd many-outlier procedure,''
  \emph{Technometrics}, vol.~25, no.~2, pp. 165--172, 1983.

\bibitem{lu2008network}
W.~Lu and A.~A. Ghorbani, ``Network anomaly detection based on wavelet
  analysis,'' \emph{EURASIP Journal on Advances in Signal Processing}, vol.
  2009, pp. 1--16, 2008.

\bibitem{mahimkar2011rapid}
A.~Mahimkar, Z.~Ge, J.~Wang, J.~Yates, Y.~Zhang, J.~Emmons, B.~Huntley, and
  M.~Stockert, ``Rapid detection of maintenance induced changes in service
  performance,'' in \emph{Proceedings of the Seventh COnference on emerging
  Networking EXperiments and Technologies}, 2011, pp. 1--12.

\bibitem{zhang2005network}
Y.~Zhang, Z.~Ge, A.~Greenberg, and M.~Roughan, ``Network anomography,'' in
  \emph{Proceedings of the 5th ACM SIGCOMM conference on Internet Measurement},
  2005, pp. 30--30.

\bibitem{xu2018unsupervised}
H.~Xu, W.~Chen, N.~Zhao, Z.~Li, J.~Bu, Z.~Li, Y.~Liu, Y.~Zhao, D.~Pei, Y.~Feng
  \emph{et~al.}, ``Unsupervised anomaly detection via variational auto-encoder
  for seasonal kpis in web applications,'' in \emph{Proceedings of the 2018
  World Wide Web Conference}, 2018, pp. 187--196.

\bibitem{laptev2015generic}
N.~Laptev, S.~Amizadeh, and I.~Flint, ``Generic and scalable framework for
  automated time-series anomaly detection,'' in \emph{Proceedings of the 21th
  ACM SIGKDD International Conference on Knowledge Discovery and Data
  Mining}.\hskip 1em plus 0.5em minus 0.4em\relax ACM, 2015, pp. 1939--1947.

\bibitem{dos2014deep}
C.~Dos~Santos and M.~Gatti, ``Deep convolutional neural networks for sentiment
  analysis of short texts,'' in \emph{Proceedings of COLING 2014, the 25th
  International Conference on Computational Linguistics: Technical Papers},
  2014, pp. 69--78.

\bibitem{chung2014empirical}
J.~Chung, C.~Gulcehre, K.~Cho, and Y.~Bengio, ``Empirical evaluation of gated
  recurrent neural networks on sequence modeling,'' \emph{arXiv preprint
  arXiv:1412.3555}, 2014.

\bibitem{kingma2013auto}
D.~P. Kingma and M.~Welling, ``Auto-encoding variational bayes,'' \emph{arXiv
  preprint arXiv:1312.6114}, 2013.

\bibitem{xu2015empirical}
B.~Xu, N.~Wang, T.~Chen, and M.~Li, ``Empirical evaluation of rectified
  activations in convolutional network,'' \emph{arXiv preprint
  arXiv:1505.00853}, 2015.

\bibitem{vaswani2017attention}
A.~Vaswani, N.~Shazeer, N.~Parmar, J.~Uszkoreit, L.~Jones, A.~N. Gomez,
  {\L}.~Kaiser, and I.~Polosukhin, ``Attention is all you need,'' in
  \emph{Advances in neural information processing systems}, 2017, pp.
  5998--6008.

\bibitem{siffer2017anomaly}
A.~Siffer, P.-A. Fouque, A.~Termier, and C.~Largouet, ``Anomaly detection in
  streams with extreme value theory,'' in \emph{Proceedings of the 23rd ACM
  SIGKDD International Conference on Knowledge Discovery and Data Mining},
  2017, pp. 1067--1075.

\bibitem{o2010nasa}
P.~O'Neill, D.~Entekhabi, E.~Njoku, and K.~Kellogg, ``The nasa soil moisture
  active passive (smap) mission: Overview,'' in \emph{2010 IEEE International
  Geoscience and Remote Sensing Symposium}.\hskip 1em plus 0.5em minus
  0.4em\relax IEEE, 2010, pp. 3236--3239.

\bibitem{jarvelin2002cumulated}
K.~J{\"a}rvelin and J.~Kek{\"a}l{\"a}inen, ``Cumulated gain-based evaluation of
  ir techniques,'' \emph{ACM Transactions on Information Systems (TOIS)},
  vol.~20, no.~4, pp. 422--446, 2002.

\end{thebibliography}
